\documentclass[runningheads]{llncs}
\usepackage{graphicx}
\usepackage{amsfonts,amssymb}
\usepackage{booktabs}
\usepackage{multirow}
\usepackage{geometry}
\begin{document}
\title{Multi-scale Feature Extraction and Fusion for Online Knowledge Distillation\thanks{This work was supported in part by the National Key R\&D Program of China (No. 2021YFB3300100), and the National Natural Science Foundation of China (No. 62171062).}}

%\institute{Beijing University of Posts and Telecommunications, Beijing, China %\and
%Springer Heidelberg, Tiergartenstr. 17, 69121 Heidelberg, Germany

\author{Panpan Zou\inst{1}\orcidID{0000-0002-7040-6242} \and
Yinglei Teng\inst{1,2}\orcidID{0000-0002-7170-4764} \and
Tao Niu\inst{1}\orcidID{0000-0001-6149-2908}}
\authorrunning{P. Zou et al.}
\institute{Beijing University of Posts and Telecommunications, Beijing, China %\and
%Springer Heidelberg, Tiergartenstr. 17, 69121 Heidelberg, Germany
\email{\{zoupanpan,lilytengtt,tasakim\}@bupt.edu.cn}\\
%\url{http://www.springer.com/gp/computer-science/lncs}
\and
Beijing Key Laboratory
of Space-ground Interconnection and Convergence
}

\maketitle              % typeset the header of the contribution

\begin{abstract}
%Online knowledge distillation conducts knowledge transfer among all student models to alleviate the reliance on pre-trained models. However, existing methods rely heavily on the prediction distributions and neglect the further exploration of the representational knowledge. In this paper, we propose a novel Multi-scale Feature Extraction and Fusion method (MFEF) for online knowledge distillation, which integrates multi-scale feature extraction and attention mechanism into a unified feature fusion framework. Our MFEF consists of three key modules to process and generate more informative feature maps: the multi-scale feature extractor including divide and concatenate operations in dimension is proposed to improve the multi-scale representation power of feature maps while the dual-attention module is designed to strengthen the important channel and spatial regions within a context adaptively. Moreover, we aggregate and fuse the former processed feature maps via a feature fusion module to assist the training of student models. Extensive experiments on CIFAR-10, CIFAR-100, and CINIC-10 show that the proposed method improves the multi-scale representation power and is more effective than other alternative methods in taking full advantage of the high-dimension feature maps.

Online knowledge distillation conducts knowledge transfer among all student models to alleviate the reliance on pre-trained models. However, existing online methods rely heavily on the prediction distributions and neglect the further exploration of the representational knowledge. In this paper, we propose a novel Multi-scale Feature Extraction and Fusion method (MFEF) for online knowledge distillation, which comprises three key components: Multi-scale Feature Extraction, Dual-attention and Feature Fusion, towards generating more informative feature maps for distillation. The multi-scale feature extraction exploiting divide-and-concatenate in channel dimension is proposed to improve the multi-scale representation ability of feature maps. To obtain more accurate information, we design a dual-attention to strengthen the important channel and spatial regions adaptively. Moreover, we aggregate and fuse the former processed feature maps via feature fusion to assist the training of student models. Extensive experiments on CIFAR-10, CIFAR-100, and CINIC-10 show that MFEF transfers more beneficial representational knowledge for distillation and outperforms alternative methods among various network architectures.

%consistently achieves superior performance against alternative methods among various network architectures.

\keywords{Knowledge distillation \and Multi-scale \and Feature fusion.}
\end{abstract}

\section{Introduction}

Driven by the advances in algorithms, computing power, and big data, deep learning has achieved remarkable breakthroughs in various vision tasks~\cite{bib1,bib2,bib3}. Increasing the network depth or width is often the key point to further improve the performance of deep neural networks. However, these models with millions of parameters demand high computational costs and huge storage requirements, making it challenging to deploy them in resource-limited or low latency scenarios. For instance, mobile phones and Internet of Things (IoT) devices. To address this problem, extensive research has been carried out to develop a lightweight model while simultaneously keeping negligible model accuracy degradation in performance. These efforts can typically be classified into network pruning, parameter quantization, low-rank approximation, and knowledge distillation.

Knowledge distillation (KD) has been demonstrated as an effective technique for model compression. The vanilla KD~\cite{bib4} method adopts a two-stage training strategy, where knowledge is transferred from the pre-trained high-capacity teacher model to a compact student model via aligning prediction distributions or feature representations~\cite{bib5}, also known as the offline distillation. Drawbacks of these methods include the fact that the high-capacity teacher is not always available, even if they are, higher computational cost and training time of the complex teacher also cannot be avoided. In addition, KD suffers from model capacity gap when the size difference is large between the student and teacher model~\cite{bib6}. 

Online knowledge distillation (OKD)~\cite{bib7,bib8,bib9,bib10} has been developed to alleviate the above issue. This paradigm is more attractive for the reason that instead of using a pre-trained high-performance teacher, it breaks the presupposed specific strong-weak relationship and simplifies the training process to an end-to-end one-stage fashion. All models are trained simultaneously by learning from each other across the training process. In the other words, knowledge is distilled and shared among all networks. Compared to the offline KD, the online KD achieves superior performance while keeping a more straightforward structure. However, popular methods concentrate on transferring logit information as soft targets in common. Although the soft targets carry richer information than one-hot labels, it is relatively unitary to make use of only the logit. Since feature maps can provide rich information about the perception, channel and spatial correlations, simply aligning or fusing cannot take full advantage of the meaningful feature representation.

In this paper, to alleviate the aforementioned limitation, we propose a novel Multi-scale Feature Extraction and Fusion method (MFEF) for online knowledge distillation, including three key components, i.e., multi-scale feature extraction, dual-attention, and feature fusion. In order to obtain more beneficial representational knowledge for distillation, we first get multi-scale features which can focus on both local details and global regions by multiple divide and concatenate operations. Then, students are guided to learn more accurate features by introducing dual-attention which boosts the representation power of important channel and spatial regions while suppressing unnecessary regions. Finally, we utilize feature fusion to integrate the acquired feature maps and feed them into a fusion classifier to assist the learning of student models.

To summarize, the main contributions of this paper are:
\vspace{-0.1cm}
\begin{itemize}
\item We propose a novel Multi-scale Feature Extraction and Fusion method (MFEF) for online knowledge distillation, which integrates the feature representation with soft targets for distillation.
\end{itemize}
\vspace{-0.1cm}
\begin{itemize}
\item We first introduce multi-scale feature extraction to improve the multi-scale representation ability of the features and provide richer information apart from simply alignment. Then the dual-attention is proposed to generate more accurate features. Furthermore, we use feature fusion to fuse the enhanced knowledge, which can improve the generalization ability for distillation.
\end{itemize}
% \vspace{-0.1cm}
\begin{itemize}
\item Extensive experiments on CIFAR-10/100~\cite{bib11} and CINIC-10~\cite{bib12} verify that the proposed MFEF can effectively enhance the multi-scale representation power of features and generate more informative knowledge for distillation.
\end{itemize}
\section{Related Work}
\vspace{-0.1cm}
Many efforts have been conducted with regard to knowledge distillation and vision tasks. In this section, we will give a comprehensive description of the related literature.
% \vspace{-0.3cm}
\subsection{Traditional Knowledge Distillation}
\vspace{-0.1cm}
The idea of transferring the knowledge from a cumbersome model to a smaller model without a significant drop in accuracy is derived from~\cite{bib13}. Traditional KD works in a two-stage fashion which needs a pre-trained teacher.~\cite{bib5} first introduces intermediate features from hidden layers, the main idea is to match the feature activations of the student and teacher model.~\cite{bib14} combines attention with distillation to further exploit more accurate information.~\cite{bib15} explores the relationships between layers by mimicking the teacher’s flow matrices using the inner product. In~\cite{bib16}, the adversarial training scheme is utilized to enable the student and teacher networks to learn the true data distribution.~\cite{bib6} introduces a teacher assistant to mitigate the capacity gap between the teacher model and student model. In~\cite{bib17}, it proposes to use the activation boundaries formed by hidden neurons for distillation. 
% \vspace{-0.3cm}

\subsection{Online Knowledge Distillation}
\vspace{-0.1cm}
Online knowledge distillation has emerged to further improve the performance of the student model and eliminate the dependency on high-capacity teacher models which are time-consuming and costly. In this paradigm, student models learn mutually by sharing the predictions throughout the training process.~\cite{bib7} is a representative method in which multiple networks work in a collaborative way. Each network imitates the peer network’s class probabilities using Kullback-Leibler divergence.~\cite{bib9} further extends DML to construct an ensemble logit as the teacher by averaging a group of students’ predictions to improve generalization ability. A fusion module is introduced to train a fusion classifier to guide the training of sub-networks in~\cite{bib10}.~\cite{bib18} adds a gate module to generate the importance score for each branch and build a stronger teacher.~\cite{bib19} proposes two-level distillation between multiple auxiliary 
\begin{figure}[t]
\begin{center}
\includegraphics[width=0.8\textwidth]{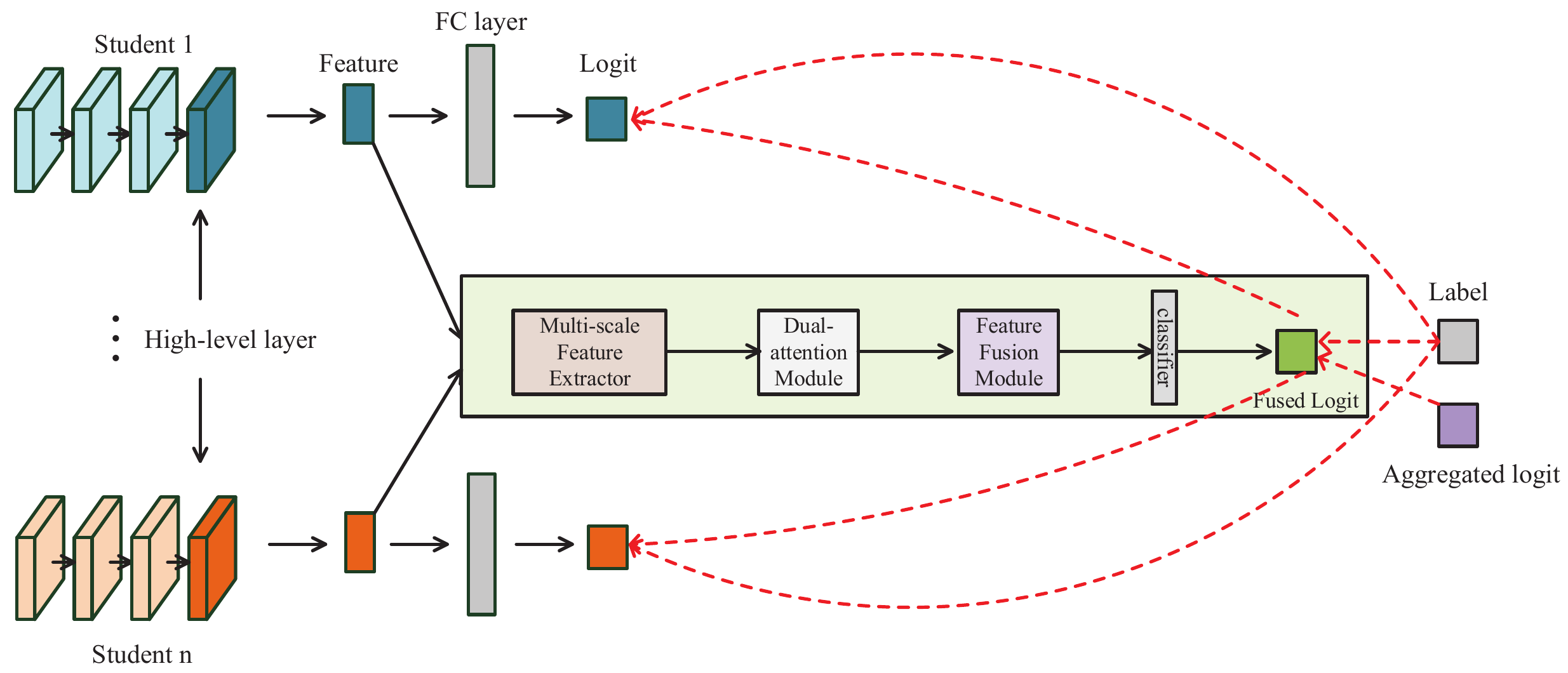}
\caption{An overview of Multi-scale Feature Extraction and Fusion (MFEF) for Online knowledge distillation. The output of high-level layer is sent to three key components (i) Multi-scale Feature Extraction: Enhance the multi-scale representation ability of feature maps. (ii) Dual-attention: Use channel and spatial attention to strengthen informative regions. (iii) Feature Fusion: Integrate knowledge among stuent models and futher improve the generalization ability.} \label{fig1}
\end{center}
\vspace{-0.8cm}
\end{figure}
peers and a group leader to enhance diversity among student models. In terms of architecture designing,~\cite{bib20} forms the student model via replacing the standard convolution with cheap convolution operations. Student and teacher models share the same networks in~\cite{bib21}, where knowledge is distilled within the network itself and knowledge from the deeper portions of the network is distilled into shallow ones.
%\vspace{-0.2cm}
\subsection{Multi-scale Feature}
\vspace{-0.1cm}
Multi-scale feature representations are of critical importance to many vision tasks. Some concurrent works focus on promoting the capability of models by utilizing multi-scale features.~\cite{bib22} constructs hierarchical residual-like connections within a residual block to represent multi-scale features at a granular level.~\cite{bib23} uses pyramidal convolution including four levels of different kernel sizes to generate multi-scale features. Similarly,~\cite{bib24} integrates information at different scales via pyramidal convolution structure for the channel-wise feature maps. A flexible and efficient hierarchical-split block is used in~\cite{bib25} to capture multi-scale features.~\cite{bib26} adopts atrous spatial pyramid pooling to probes convolutional features on multiple scales for semantic image segmentation.
\vspace{-0.2cm}
\section{Proposed Method}
In this section, we describe the framework and loss function in detail. An overview of MFEF is illustrated in Fig.~\ref{fig1}. Different from the existing KD methods, MFEF digs deeper into the information provided by feature maps including multi-scale representation ability and the channel and spatial attention.
\vspace{-0.3cm}

\subsection{Problem Definition}
The key idea of knowledge distillation is that soft targets contain the dark knowledge which can be used as a supervisor to transfer knowledge to the student model. Given a labeled dataset $D\{x_i,y_i\}_{i=1}^N$, with $N$ samples, $x_i$ is the $i$th input sample and $y_i\in \{1,2,...,M\}$ is the corresponding ground-truth label. $M$ is the total number of classes in the dataset. Consider $n$ student models $\{S_j\}_{j=1}^n$, the logit produced by the last fully connected layer of the student $S_j$ is denoted as $z_j=\{z_j^1,z_j^2,...,z_j^M\}$. Then the probability of the $j$th student for the sample $x_i$ over the $m$th class $p_j^m(x_i)$ can be estimated by a softmax function,

\begin{equation}
 p_j^m(x_i)=\frac{\exp(z_j^m/T)}{ {\textstyle \sum_{m=1}^{M}\exp (z_j^m/T)} },\label{eq1}
\end{equation}
where $T$ is the temperature which produces a more softened probability distribution as it increases. Specifically, when $T=1$, it is defined as the original softmax output, we consider writing it as $p_j^m(x_i)$; otherwise it is rewritten as $\tilde{p} _j^m(x_i)$. For multi-class classification, the objective is to minimize the cross-entropy loss between the softmax outputs and the ground-truth labels,
\begin{equation}
L_j^{CE}=-\sum_{i=1}^{N}\sum_{m=1}^{M}l_i\log(p_j^m(x_i)),\label{eq2}
\end{equation}
where $l_i=1$ if $y_i=m$, and 0 otherwise.
\begin{figure}[t]
\begin{center}
\includegraphics[width=1\textwidth]{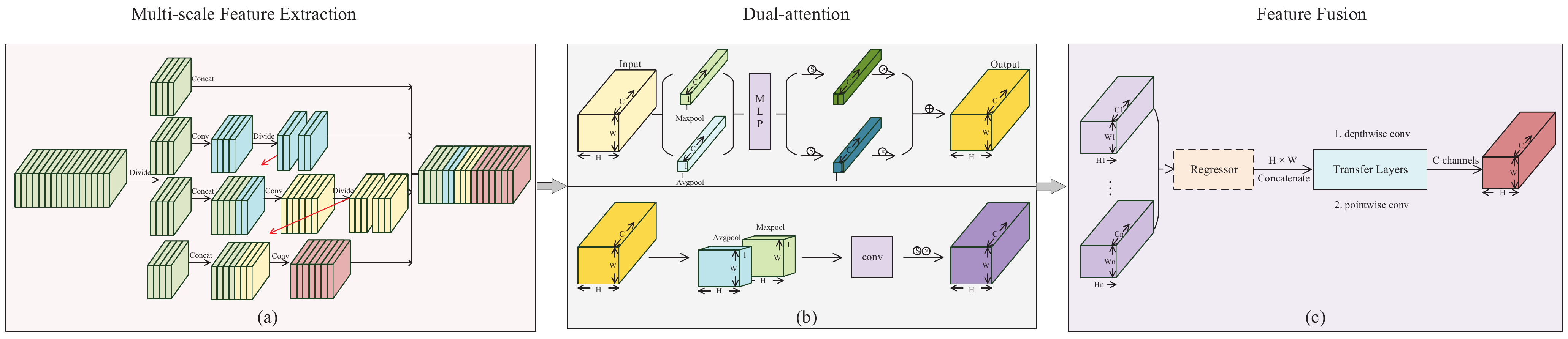}
\caption{The structure of key components: (a). Multi-scale feature extraction. (b) Dual-attention. (c) Feature Fusion} \label{fig2}
\end{center}
\vspace{-0.6cm}
\end{figure}
Knowledge transfer is facilitated by matching the softened probability of the student model $\tilde{p} _j^m(x_i)$ and the teacher model  $\tilde{p} _t^m(x_i)$. We introduce the distillation loss of $j-th$ student model in the form of Kullback-Leibler Divergence

\begin{equation}
L_j^D=\sum_{i=1}^{N}\sum_{j=1}^{M}\tilde{p}_t^m(x_i)\log\frac{\tilde{p}_t^m(x_i)}{\tilde{p}_j^m(x_i)}.\label{eq4}
\end{equation}

\subsection{MFEF Framework}
From a global perspective, the main idea of MFEF is to enhance the multi-scale representation power of feature maps and generate more informative knowledge for distillation. The details of each key component are explained in the following.
\vspace{-0.4cm}
\subsubsection{Multi-scale Feature Extraction.} Aligning the soft targets of teacher and student models enhances the model generalization, but it ignores the feature maps which contain rich information. In addition to the soft targets, inspired by~\cite{bib25}, we introduce multi-scale feature extraction to generate multi-scale features which are of significant importance for vision tasks. As shown in Fig.~\ref{fig2} (a), the extraction includes multiple divide and concatenate operations in the channel dimension to enhance the information flow between different groups. We use the feature maps of the last layer as the input for the reason it has high-level semantic information which is richer and specific. For the convenience of notation, we name the feature map of the $j$th student model as $F_j$ and the extraction as $E$. $D$ and $C$ represent the divide and concatenate operations, respectively. First, $F_j$ is divided into $p$ groups $\{F_{j1},F_{j2},...,F_{jp}\}$. The first group $F_{j1}$ is output straightforward and the second part is sent to a convolution operation and then is divided into two sub-groups $D_{21}$ and $D_{22}$. One of them is exported to the output and the other is concatenated with the next part. The rest other than the last group follows the concatenate-convolution-export-divide procedure. The last part does not need the divide operation. We define the output as
\begin{equation}
E(F_j)=C(F_{j1},D_{22},D_{32},...,Conv(C(F_{jp},D_{p-1,2})).\label{eq5}
\end{equation}

The multi-scale feature extraction can generate feature maps that contain multiple scales of receptive fields. The more features are concatenated, the larger the receptive field is. Larger receptive fields can capture global information while the smaller ones can focus on details. Such a combination can generate more meaningful feature maps to improve the performance of distillation.

%\begin{figure}
%\begin{center}
%\includegraphics[width=0.8\textwidth]{fig3.png}
%\caption{The architecture of a Dual-attention Module. Max-pooling and average-pooling are used to obtain finer channel and spatial attention.} \label{fig3}
%\end{center}
%\vspace{-0.6cm}
%\end{figure}
\vspace{-0.4cm}
\subsubsection{Dual-attention.} After the extraction, we utilize dual-attention to dig deeper into the feature maps (see Fig.~\ref{fig2} (b)). Channel and spatial attention focus on “what” and “where” are important, and we apply them in a sequential manner. We denote the multi-scale feature map $E(F_j)\in \mathbb{R} ^{C\times H\times W}$ as the input, where $C$, $H$, $W$ represent its channel numbers, height, and width, respectively. Average-pooling and max-pooling are used in combination to obtain finer attention. 

For channel attention, we denote $a_c,m_c\in \mathbb{R}^{C\times 1\times 1}$ as the vectors after average-pooling and max-pooling. The weight $w_c\in \mathbb{R}^{C\times 1\times 1}$ of channel is
\begin{equation}
w_c=\sigma (W(a_c))) + (W(m_c))),\label{eq6}
\end{equation}

where the symbol $\sigma$ denotes the Sigmoid function, $W$ is the weight of a multi-layer perceptron (MLP). The channel attention output $AT_j^c$ is
\begin{equation}
AT_j^c=w_c\otimes F_M,\label{eq7}
\end{equation}
where $\otimes$ refers to element-wise multiplication. Similarly, we denote the average-pooling and max-pooling vector  $a_s,m_s\in \mathbb{R}^{1\times H\times W}$, $w_s\in \mathbb{R}^{1\times H\times W}$ is
\begin{equation}
w_s=\sigma (conv(a_s;m_s)),\label{eq8}
\end{equation}
where $conv$ represents a convolution operation. The output $AT_j^s$ is
\begin{equation}
AT_j^s=w_s\otimes AT_j^c,\label{eq9}
\end{equation}

Dual-attention can strengthen the informative channel and spatial regions while suppressing the less important ones and thus generate more informative outputs which can focus on useful regions within a context adaptively.
%\begin{figure}[h]
%\begin{center}
%\includegraphics[width=0.6\textwidth]{fig4.png}
%\caption{The architecture of a Dual-attention Module. Max-pooling and average-pooling are used to obtain finer channel and spatial attention.} \label{fig4}
%\end{center}
%\vspace{-0.6cm}
%\end{figure}
%\vspace{-0.6cm}
\subsubsection{Feature Fusion.} We propose feature fusion to aggregate and maximize the usage of the student models’ information. The structure of it is illustrated in Fig.~\ref{fig2} (c). Specifically, we first concatenate the meaningful feature maps of students that have been processed previously, i.e., $\{AT_1^s,AT_2^s,...,AT_j^s\}$. If the resolutions of the feature maps are different, we apply a convolutional regressor to make them identical. Then we concatenate them and sent the results to the transfer layers which consist of a sequence of depthwise and pointwise convolution operations. Finally, we fuse the student models’ feature information and feed it into a fusion classifier which is supervised by ground truth labels.
\vspace{-0.3cm}
\subsection{Loss Function}
% \vspace{-0.3cm}
The cross-entropy loss of the $j$th student and the fused classifier is $L_j^{CE}$ and $L_f^{CE}$, respectively, as described in Eq. (2). We further define the aggregated logit of students as $z_a^m=\frac{1}{n}\sum_{j=1}^{n}z_j^m$ and probability as $p_a^m$. The fusion classifier is trained with KL divergence
\begin{equation}
L_a^D=L_a^{KL}(\tilde {p}_a^m,\tilde {p}_f^m),\label{eq11}
\end{equation}
This loss is used to transfer the knowledge of the student models to the fusion classifier. Then the fusion classifier facilitate the knowledge which contains informative feature representations transferring back to the student models via minimizing the distillation loss
\begin{equation}
L_f^D=\sum_{j=1}^{n} L_f^{KL}(\tilde {p}_f^m,\tilde {p}_j^m),\label{eq12}
\end{equation}

Finally, we derive the total training objective as 
\begin{equation}
L_{total}=L_{CE}+T^2L_D.\label{eq14}
\end{equation}

where $L_{CE}$ is the sum of cross-entropy of students and fused classifier. $L_D$ refers to the sum of $L_a^D$ and $L_f^D$. Because the gradients produced by the soft targets are scaled by $1/{T^2}$, thus $L_D$ is multiplied with $T^2$ to keep the contributions of $L_{CE}$ and $L_D$ roughly balanced.
% \vspace{-0.1cm}
\section{Experiment}

In this section, we conduct comprehensive experiments to evaluate the performance of MFEF on three datasets and various widely-used neural networks. We choose various related methods under different settings for comparison and show the results to demonstrate that MFEF generalizes well among different numbers and types of models. Finally, evaluation of each component are carried out.

\begin{table}
\centering
\caption{Comparisons with closely related methods on CIFAR 10 and CIFAR-100 with seven different networks. Top-1 error rates$(\%)$ are reported. Two same student models are used for each method. FFL-S and MFEF-S refer to the results of the student model, and FFL and MFEF refer the results of fused classifiers.}
\label{tab1}
\begin{tabular}{c|cccccccc} 
\toprule
\begin{tabular}[c]{@{}c@{}}\\Dataset\end{tabular} & Network       & Baseline & KD    & DML   & FFL-S & FFL   & MFEF-S          & MFEF             \\ 
\midrule
\multirow{7}{*}{CIFAR-10}                         & ResNet-20     & 7.32     & 7.18  & 6.63  & 6.49  & 6.22  & \textbf{6.38 }  & \textbf{6.08 }   \\
                                                  & ResNet-32     & 6.77     & 6.69  & 6.52  & 6.06  & 5.78  & \textbf{5.59 }  & \textbf{5.41 }   \\
                                                  & ResNet-56     & 6.30     & 6.14  & 5.82  & 5.46  & 5.26  & \textbf{5.28 }  & \textbf{4.82 }   \\
                                                  & ResNet-110    & 5.64     & 5.47  & 5.21  & 5.18  & 4.83  & \textbf{4.81 }  & \textbf{4.52 }   \\
                                                  & WRN-16-2      & 6.78     & 6.40  & 5.49  & 6.09  & 5.97  & \textbf{5.33 }  & \textbf{4.99 }   \\
                                                  & WRN-40-2      & 5.34     & 5.24  & 4.72  & 4.75  & 4.60  & \textbf{4.51 }  & \textbf{4.02 }   \\
                                                  & DenseNet40-12 & 6.87     & 6.81  & 6.50  & 6.72  & 6.24  & \textbf{5.79 }  & \textbf{5.30 }   \\ 
\midrule
\multirow{7}{*}{CIFAR-100}                         & ResNet-20     & 31.08    & 29.94 & 29.61 & 28.56 & 26.87 & \textbf{28.46 } & \textbf{26.30 }  \\
                                                  & ResNet-32     & 30.34    & 29.82 & 26.89 & 27.06 & 25.56 & \textbf{26.36 } & \textbf{24.84 }  \\
                                                  & ResNet-56     & 29.31    & 28.61 & 25.51 & 24.85 & 23.53 & \textbf{24.22 } & \textbf{23.15 }  \\
                                                  & ResNet-110    & 26.30    & 25.67 & 24.49 & 23.95 & 22.79 & \textbf{23.37 } & \textbf{22.16 }  \\
                                                  & WRN-16-2      & 27.74    & 26.78 & 26.16 & 25.72 & 24.74 & \textbf{24.66 } & \textbf{22.93 }  \\
                                                  & WRN-40-2      & 25.13    & 24.43 & 22.77 & 22.06 & 21.05 & \textbf{21.76 } & \textbf{20.60 }  \\
                                                  & DenseNet40-12 & 28.97    & 28.74 & 26.94 & 27.21 & 24.76 & \textbf{26.81 } & \textbf{24.27 }  \\
\bottomrule
\end{tabular}
\end{table}
\vspace{-0.3cm}
\subsection{Experiment Settings}
\subsubsection{Datasets and Architecture.} We incorporate three image classification datasets in the following evaluations. (1) CIFAR-10 which contains 60000 colored natural images (50000 training samples and 10000 test samples) over 10 classes. (2) CIFAR-100 consists of 60000 images (50000 training samples and 10000 test samples) drawn from 100 classes. (3) CINIC-10 consists of images from both CIFAR and ImageNet. It is more challenging than CIFAR-10. It contains 90000 train samples and 90000 test samples. For CIFAR-10/100, there are seven popular networks used, namely ResNet-20, ResNet-32, ResNet-56, ResNet-110, WRN-16-2, WRN-40-2, and DenseNet-40-12. For CINIC-10, we use MobileNetV2 and ResNet-18 following the settings in~\cite{bib12}.
\vspace{-0.3cm}
\subsubsection{Settings.} We apply horizontal flips and random crop from an image padded by 4 pixels for data augmentation in training. We use SGD as the optimizer with Nesterov momentum 0.9, weight decay of 1e-4 for student models and 1e-5 for fusion and mini-batch size of 128. The models are trained for 300 epochs for all datasets. We set the initial learning rate to 0.1 and is multiplied by {0.1} at {150, 225} epochs. We set the temperature $T$ to 3 empirically and $\alpha = 80$ for ramp-up weighting. For the case of student models have same architecture, the low-level layers are shared following~\cite{bib19}. When output channels of the feature maps are different, the feature fusion is designed to match the smaller one. For fair comparison, we set the number of student models to two. The top-1 error rate $(\%)$ of the best student over 3 runs is reported.

\subsection{Experiment Results}
\subsubsection{Results on CIFAR-10/100.} As shown in Table ~\ref{tab1}, we evaluate the effectiveness of MFEF on CIFAR-10 and CIFAR-100 based on several popular networks. Since our goal is to distill more powerful feature representations for online distillation, we compare MFEF with the offline KD, logit-only online method DML, and fusion-only method FFL. For the offline KD, it employs a pre-trained ResNet-110 as the teacher model. For DML, we report the top-1 error rate of the best student. FFL-S and MFEF-S represent the results of the best student and FFL and MFEF indicate the results of the fused classifier. 

The results clearly show the performance advantages of our MFEF. Specifically, MFEF improves by approximately $1\%$ and $2\%$ of the backbone networks. MFEF also achieves the best top-1 error rate compared with the closely related online distillation methods. For instance, on CIFAR-10, MFEF-S achieves lower error rates than FFL-S by approximately $0.5\%$, $0.8\%$, and $1\%$ on ResNet-32, WRN-16-2, and DensNet-40-12, respectively. MFEF improves FFL by about $1\%$ on WRN-16-2 and DensNet-40-12; While on CIFAR-100, MFEF achieves $0.6\%$, $0.7\%$, and $1\%$ increase on ResNet-56, ResNet -32, and WRN-16-2, respectively. MFEF is higher by about $1.8\%$ on WRN-16-2 compared with FFL. These improvements attribute to the integration of the multi-scale feature extraction and the attention mechanism and the feature fusion of student models.

\begin{figure}[t]
\centering
\includegraphics[width=0.7\textwidth]{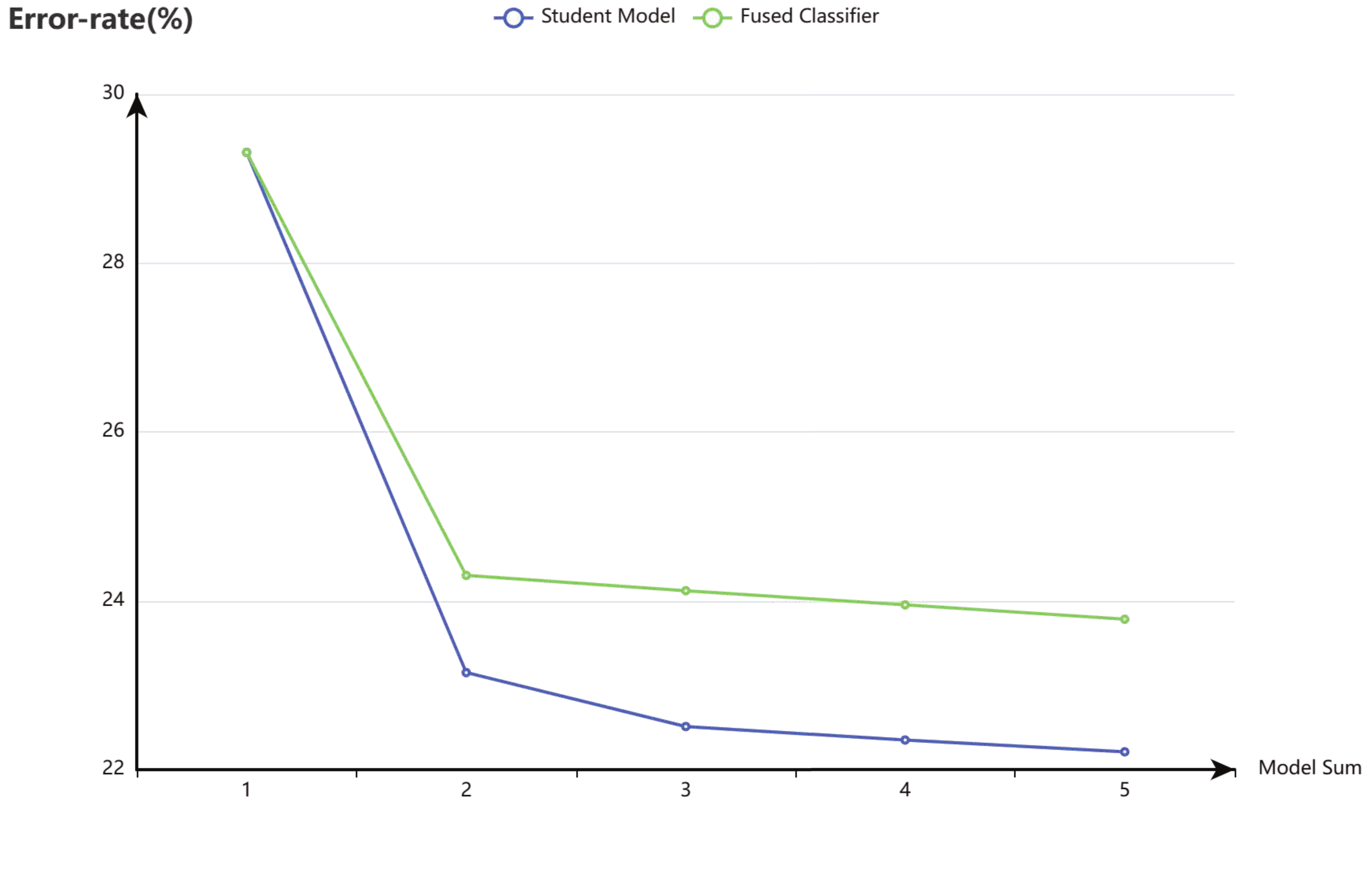}
\caption{Evaluating the impact of expansion of student models on CIFAR-100 using ResNet-56.}\label{fig3}
\vspace{-0.3cm}
\end{figure} 

\begin{table}
\centering
\caption{Top-1 error rate $(\%)$ comparison with FFL on CINIC-10.}
\label{tab2}
\begin{tabular}{cccccc} 
\toprule
\begin{tabular}[c]{@{}c@{}}\\Network\end{tabular} & Baseline & FFL-S & FFL   & MFEF-S          & MFEF             \\ 
\midrule
MobileNetV2                                       & 18.07    & 17.85 & 16.10 & \textbf{17.56 } & \textbf{15.66 }  \\
ResNet-18                                         & 13.94    & 13.33 & 12.67 & \textbf{13.22 } & \textbf{12.39 }  \\
\bottomrule
\end{tabular}
\vspace{-1cm}
\end{table}

\subsubsection{Results on CINIC-10.} In this section, we compare the top-1 error rate of MFEF with FFL based on MobileNetV2 and ResNet-18. As shown in Table ~\ref{tab2}, FFL and MFEF both reduces the error rate of the baseline and MFEF shows higher improvement of performance in both student models and fused classifier. In case of MobileNetV2, MFEF improves by around $0.5\%$, $0.3\%$, and $0.4\%$ compared to the baseline and FFL for student model and fused classifier. Based on these experiments, we could confirm that thanks to the enhancement of the multi-scale representation power, higher-quality knowledge is transferred among all student models and consequently achieves a lower error rate than others.
\vspace{-0.2cm}
% \begin{table}[]
% \begin{center}
% \caption{Top-1 error rate $(\%)$ comparisons with other online distillation methods for training three students model on CIFAR-100. ONE and ONE-E refer to the results of the student models and the gated ensemble teacher. }\label{tab3}%
% \begin{tabular}{lll}
% \toprule
% Method & ResNet-32 & ResNet-56 \\ \midrule
% ONE    & 26.64    & 24.63    \\
% FFL-S  & 26.30     & 24.51    \\
% MFEF-S  & 26.04    & 24.12    \\
% ONE-E  & 24.75    & 23.27    \\
% FFL    & 24.31    & 23.20    \\
% MFEF    & 24.03     & 22.51    \\
% \bottomrule
% \end{tabular}
% \end{center}
% \vspace{-0.6cm}
% \end{table}

\begin{table}
\centering
\caption{Top-1 error rate $(\%)$ comparisons with other online distillation methods for training three students model on CIFAR-100. ONE and ONE-E refer to the results of the student models and the gated ensemble teacher. }
\label{tab3}
\begin{tabular}{ccc} 
\toprule
\begin{tabular}[c]{@{}c@{}}\\Method\end{tabular} & ResNet-32       & ResNet-56        \\ 
\midrule
ONE                                              & 26.64           & 24.63            \\
FFL-S                                            & 26.30           & 24.51            \\
MFEF-S                                           & \textbf{26.04 } & \textbf{24.12 }  \\
ONE-E                                            & 24.75           & 23.27            \\
FFL                                              & 24.31           & 23.20            \\
MFEF                                             & \textbf{24.03 } & \textbf{22.51 }  \\
\bottomrule
\end{tabular}
\end{table}
\begin{table}
\centering
\caption{Top-1 error rate $(\%)$ comparisons with other online distillation methods for different architectures of student models on CIFAR-100.}
\label{tab4}
\begin{tabular}{ccc|cc} 
\toprule
\multirow{2}{*}{\begin{tabular}[c]{@{}c@{}}\\~\\\end{tabular}} & Net1            & \multicolumn{1}{c}{Net2}     & Net1           & Net2             \\
                                                               & ResNet-32       & \multicolumn{1}{c}{WRN-16-2} & ResNet-56      & WRN-40-2         \\ 
\midrule
DML                                                            & 28.31           & 26.45                        & 26.75          & 23.33            \\
FFL                                                            & 27.06           & 25.93                        & 26.23          & 23.06            \\
MFEF                                                           & \textbf{26.38 } & \textbf{25.16 }              & \textbf{25.7 } & \textbf{22.39 }  \\
\bottomrule
\end{tabular}
\end{table}
\subsubsection{Expansion of Student Models.} The impact of increasing the number of student models is illustrated in Fig.\ref{fig3}. We conduct experiments on ResNet-56. Not surprisingly, the performance of both students and the fusion classifier improves as the number of student models increases. When the student models expanded to 3, MFEF still performs competitively against ONE and FFL, as shown in Table \ref{tab3}. We can see that MFEF-S achieves an approximately $0.3\%$ and $0.6\%$ performance improvement on ResNet-32 compared to FFL-S and ONE-S, respectively. The fusion classifier yields an about $0.7\%$ and $0.8\%$ improvement on ResNet-56 superior to ONE and FFL.

\vspace{-0.3cm}
\subsubsection{Different Architecture.} To verify the generalization of MFEF on different model architectures, we conduct experiments on ResNet and WRN in Table \ref{tab4}. We set ResNet as Net1 and WRN as Net2. MFEF shows better performance than DML and FFL in both Net1 and Net2. An interesting observation is that when MFEF is applied, the smaller network (Net1) improves significantly compared to the larger one. For example, when compared with DML, MFEF is higher by about $2\%$ and $1.3\%$ on ResNet-32 and WRN-16-2. This is because MFEF can aggregate and fuse all networks’ feature maps and transfer the informative knowledge of the larger network to the smaller one better. 
\vspace{-0.3cm}
\begin{table}[]
\begin{center}
\caption{Evaluating the effectiveness of each component on CIFAR-100 using ResNet-110.}\label{tab5}%
\begin{tabular}{l|l|ll}
\toprule
Case & Component            & Student & Fused \\ \midrule
A    & Backbone             &26.30    & -     \\
B    & Backbone+MSFE        & 24.35   & -     \\
C    & Backbone+OKD         & 24.79   & -     \\
D    & Backbone+MSFE+OKD    & 23.54   & 22.54 \\
E    & Backbone+MSFE+DA+OKD & 23.37   & 22.16 \\
\bottomrule
\end{tabular}
\end{center}
\vspace{-1cm}
\end{table}
\subsubsection{Component Effectiveness Evaluation.} To further validate the benefit of each component, we conduct various ablation studies on CIFAR-100 on ResNet-110. Specifically, we perform experiments in five cases of ablations. As shown in Table \ref{tab5}, Case A refers to the model trained from scratch. Case B and C refer to the network where only the multi-scale feature extraction (MSFE) and OKD are included. And they improved by around $2\%$ and $1.5\%$ compared to the backbone. When both MSFE and OKD are applied in Case D, the student model achieves a higher accuracy by around $2.8\%$ compared to Case A. When we get rid of MSFE from Case D (Case C), the performance decrease sharply by about $1.3\%$, which confirms the usefulness of the MSFE. Dual-attention (DA) is added in Case E based on Case D. This increases the performance by around $0.2\%$ and $0.4\%$ of the student models and the fused classifier, respectively, and it has more influence on the fused classifier. The improvements manifest that MSFE has a more significant impact on the model performance, which is mainly attributed to the enhancement of the multi-scale representation ability.
\vspace{-0.3cm}
\section{Conclusion}
We present a novel Multi-scale Feature Extraction and Fusion method (MFEF) for online knowledge distillation . It integrates multi-scale extraction and attention mechanism into a unified feature fusion framework. Different from existing online knowledge distillation methods, we enhance the multi-scale representation ability of the feature maps and then fuse them from student models to assist the training process by transferring more informative knowledge. Extensive experiments based on three datasets show the superiority of our method compared to prior works. Results on various networks also demonstrate that the proposed method can be broadly applied to a variety of architectures from a very small scale to a large one.
\vspace{-0.3cm}

\nocite{*}
\bibliography{references} %bibfile_name
\bibliographystyle{splncs04}
\end{document}